\documentclass[10pt]{article}
\usepackage[a4paper,margin=1.7cm]{geometry}
\usepackage[utf8]{inputenc}
\usepackage[T1]{fontenc}
\usepackage{lmodern}
\usepackage{microtype}
\usepackage{amsmath,amssymb}
\usepackage{bm}
\usepackage{graphicx}
\usepackage{booktabs}
\usepackage{tabularx}
\usepackage[numbers,sort]{natbib}
\usepackage[hidelinks]{hyperref}
\usepackage{url}
\graphicspath{{figures/}}
\newcommand{\q}{\mathbf{q}}
\newcommand{\f}{\mathbf{f}}

\title{\vspace{-2.4em}A four-view smartphone video dataset of a robot-held
flexible strip, with camera poses and a 3D LaGSplat baseline\vspace{-0.4em}}
\author{Louen Pottier\\
\small Université Paris-Saclay, CEA, List, F-91120 Palaiseau, France\quad
\texttt{louen.pottier@cea.fr}}
\date{\small October 2026\vspace{-0.6em}}

\begin{document}
\maketitle

This note describes a multi-view video dataset of a deformable object driven
by a measured input, with the measured pose of every camera, and a first 3D
result obtained on it with LaGSplat~\cite{pottier2026lagsplat}. Please cite this
note when using the dataset. LaGSplat learns, from video, a low-dimensional
latent state $\q$ that plays two roles at once: it is the generalized coordinate
of a learned dissipative Lagrangian, and the conditioning variable of a Gaussian
Splatting decoder whose primitives are explicit points
$\mu_i(\q)$ that move with the object. A force $\f$ applied on the
reconstructed object then enters the equations of motion as the generalized
force $J(\q)^\top \f$, with $J=\partial\mu/\partial\q$. Fitting the decoder in
3D calls for synchronized multi-view footage. In~\cite{pottier2026lagsplat},
the 3D experiments use synthetic views rendered by
SHARP; this dataset and the results below are not
part of it.

\begin{figure}[t]
  \centering
  \includegraphics[width=0.86\textwidth]{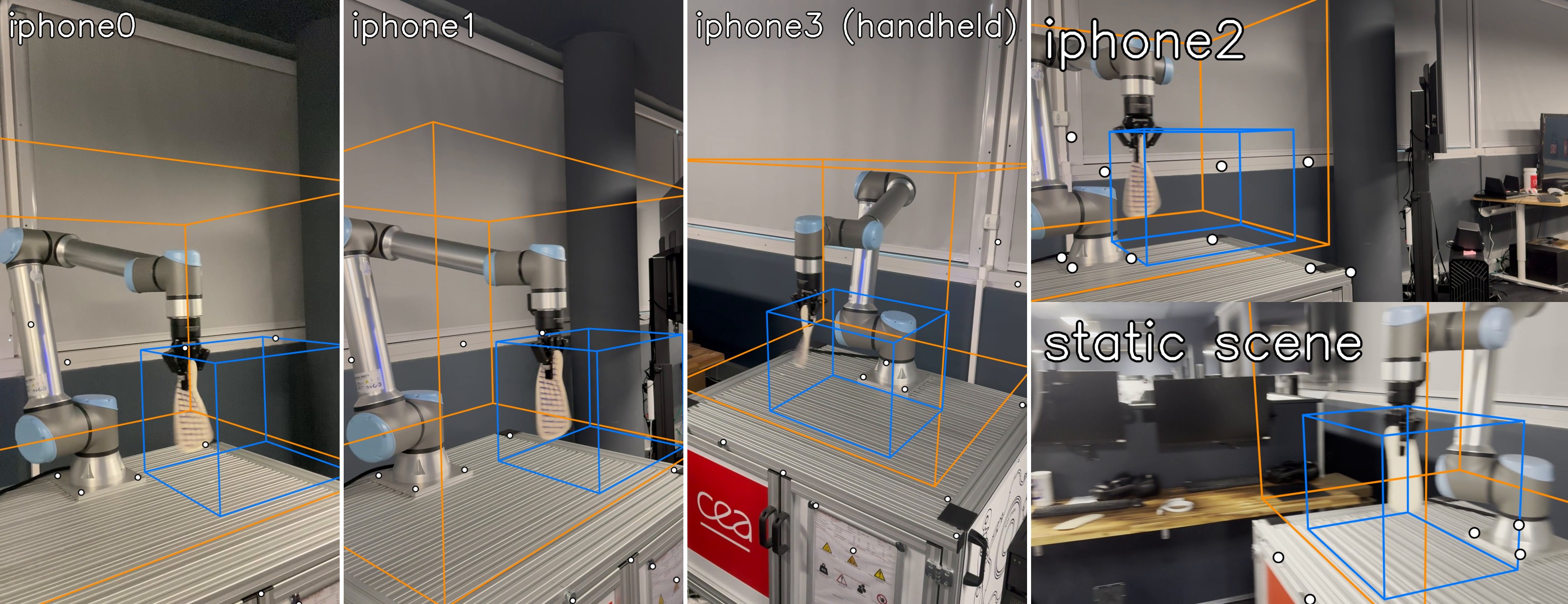}
  \caption{The four synchronized views at the same frame $k$, and one frame of
  the static scene. The zone $s$ (orange), the zone $q$ (blue) and the static
  landmarks (white) are projected with the measured pose of each camera.}
  \label{fig:views}
\end{figure}

\paragraph{Setup.}
A robot arm holds a flexible strip, about 23\,cm long, in a gripper. The end effector moves by jerks along a nearly one-dimensional
horizontal path; after each jerk the strip oscillates freely and its motion
decays. Four iPhones film the scene at $1920\times1080$ and $60000/1001$\,fps.
Three of them (iphone0, iphone1, iphone2) are fixed and mounted
rigidly to one another; iphone0 is a different model from the other three.
The fourth, iphone3, is handheld and moves around the scene for the first
seven minutes, then is held nearly still. In a separate session, with every
object in the room still, robot included, iphone3 filmed the same room in two clips,
which give a static reconstruction of the room
(Fig.~\ref{fig:views}).

\paragraph{Synchronization.}
The coarse offsets come from a clap and from the motion of the arm, whose jerks
form an irregular, hence unambiguous, sequence of instants. The clock rates,
read on the container timestamps, are not equal: over the recording, iphone0
drifts by $9$ frames with respect to the three others. All four videos are therefore resampled on the clock of iphone1:
frame $k$ of every video shows the same instant, taken as the nearest source
frame, with a residual of at most half a frame (8\,ms). Each video has $36\,734$ frames ($612.8$\,s).

\paragraph{Camera geometry.}
No calibration target was filmed; all poses come from the scene. The world
frame is the camera frame of iphone0 (OpenCV convention), and the unit of
length is the distance between iphone0 and iphone1 (about $0.38$\,m).
\begin{itemize}\setlength{\itemsep}{0pt}
\item Fixed cameras: focal lengths, radial distortion and poses are solved by a
joint bundle adjustment on manually clicked correspondences. To check the result, points are
triangulated from iphone0 and iphone1 only, then projected into iphone2: they
fall $3$ to $6$\,px from where they appear in its image.
\item Handheld camera: one pose per frame, by PnP against a SIFT
map built from iphone3's own frames. Poses with too few inliers or with
non-physical jumps are removed: $25\,381$ of the $36\,734$ frames have a
pose.
\item Static clips: structure from motion with
COLMAP on key frames, aligned to the world frame by a
similarity. A block of poses had been placed on the wrong side of a nearly symmetric cabinet; these
are corrected and flagged.
\end{itemize}

\paragraph{State and zones.}
The motion of the arm has a single parameter. The file \texttt{state.csv} gives
it per frame as $s(t)$, the arc length of the gripper along its path, tracked
in iphone1; it is the measured input of the system. The same file gives the
latent state $\q(t)\in\mathbb{R}^2$ of the strip, as encoded by the baseline
below, for reference. The file \texttt{zones.json} gives two oriented boxes in
the world frame (Fig.~\ref{fig:views}): the zone $q$ contains the strip, the
zone $s$ the moving part of the arm.

\begin{table}[t]
\centering\footnotesize
\begin{tabularx}{\textwidth}{@{}lX@{}}
\toprule
File & Content \\
\midrule
\texttt{scene\_mobile/iphone\{0,1,2,3\}.mp4} & synchronized session, H.264, 36\,734 frames each, no audio \\
\texttt{scene\_mobile/frames.csv} & per frame $k$: common time, source frame of each camera, residual offset \\
\texttt{scene\_mobile/sync.json} & cut, clock rates, dropped frames, checks \\
\texttt{scene\_statique/iphone3\_fixed\_scene\{,\_2\}.mp4} & static room, $1920\times1080$, 30\,fps, 6\,303 and 1\,722 frames \\
\texttt{geometry/cameras\_fixed.json} & intrinsics, distortion and pose of iphone0, 1, 2 \\
\texttt{geometry/iphone3\_poses.csv} & pose of iphone3 per frame $k$ (with \texttt{iphone3\_intrinsics.json}) \\
\texttt{geometry/static\_poses.json} & poses of the static clips, in the same world frame \\
\texttt{geometry/landmarks.json}, \texttt{zones.json} & 3D static landmarks; zones $s$ and $q$ \\
\texttt{geometry/state.csv} & per frame $k$: input $s(t)$ and baseline latent state $\q(t)$ \\
\bottomrule
\end{tabularx}
\caption{Files of the dataset.}
\label{tab:files}
\end{table}

\paragraph{Baseline.}
We fit LaGSplat in 3D on this dataset, with three changes
to~\cite{pottier2026lagsplat}.
(i) The arm is an input: the decoder is conditioned on $(s,\q)$ and the
dynamics is $M\ddot\q=-\nabla V(\q)-C\dot\q+G\ddot s$, with constant mass $M$
and dissipation $C$ and an invex potential $V$, which has a single minimum. The
acceleration of the gripper enters as a forcing term, in the same way as the
pressure forcing of~\cite{pottier2026lagsplat}.
(ii) The affine decoder is replaced by a shared deformation network that
outputs the displacement, scale, rotation and colour of each Gaussian. $J$ is
no longer constant and is computed at the current state; the force transport
$J^\top\f$ is unchanged, since it needs explicit moving primitives, not a
linear decoder.
(iii) The background is a static reconstruction of the room from the static
clips; only the Gaussians in the zone $q$ depend on $(s,\q)$, those in the zone
$s$ on $s$ alone. The model has about $10^5$ Gaussians, and the camera poses
are refined during training.
On the last $60$\,s, held out, the rendered images (960\,px) reach a PSNR of
$31.4$ to $32.8$\,dB for the fixed cameras and $26.5$\,dB for the handheld one;
restricted to the projection of the zone $q$, which contains the strip, $28.4$
to $30.0$\,dB and $26.2$\,dB. Held-out images of the static
scene reach $24.7$\,dB, after a gain and bias per colour channel.
For the dynamics, rollouts are started from the measured state on the held-out
part; their RMS error, divided by the standard deviation of $\q$, is $0.53$ over
$0.25$\,s and $0.83$ over $2$\,s. The model runs as a real-time 3D demo in a web
browser, in which the strip responds to forces applied with the mouse:
\url{https://louenpottier.github.io/xrdays.html}.

\paragraph{Limitations.}
The calibration error is a few pixels, not sub-pixel, and some poses of
iphone3 are wrong during the handheld part. $s(t)$ is read in one view, in
pixels. The colours of iphone0 differ from the other cameras.

\paragraph{Availability.}
Dataset and checkpoints: \href{https://doi.org/10.5281/zenodo.23601189}{\texttt{10.5281/zenodo.23601189}}, licence CC BY-NC 4.0.\\
Training code: \url{https://github.com/LouenPottier/lagsplat-mpov}.

{\footnotesize
\bibliographystyle{unsrtnat}
\bibliography{refs}
}

\end{document}